\expandafter\def\csname ver@pdfx.sty\endcsname{9999/12/31 v999 pdfx disabled}
\expandafter\def\csname opt@pdfx.sty\endcsname{useBOM,pdf17,a-1a}

\providecommand{\hypersetup}[1]{}
\providecommand{\pdfstringdefDisableCommands}[1]{}
\providecommand{\urlstyle}[1]{}

\documentclass{ceurart}

\usepackage{graphicx}
\usepackage{todonotes}
\usepackage{hyperref}
\usepackage{svg}
\usepackage{amsmath}
\usepackage{adjustbox}
\usepackage{cleveref}
\usepackage{cancel}
\usepackage{multirow}

\usepackage[compatibility=false]{caption}  

\usepackage{tikz}
\usepackage{pgfplots}
\pgfplotsset{compat=1.9}
\usepackage{color}

\begin{document}

\copyrightyear{2024}
\copyrightclause{Copyright for this paper by its authors.
  Use permitted under Creative Commons License Attribution 4.0
  International (CC BY 4.0).}

\conference{LLM+KG'24: The International Workshop on Data Management Opportunities in Unifying Large Language Models + Knowledge Graphs,
  August 25, 2024, Guangzhou, China}

\title{From Instructions to $\mathit{ODRL}$ Usage Policies: An Ontology Guided Approach}

\author[1]{Daham M. Mustafa}[%
email=daham.mohammed.mustafa@fit.fraunhofer.de,
]
\cormark[1]

\author[1]{Abhishek Nadgeri}

\author[1,3]{Diego Collarana}

\author[1,2]{Benedikt T. Arnold}

\author[2]{Christoph Quix}

\author[1,2]{Christoph Lange}

\author[1,2]{Stefan Decker}

\address[1]{Fraunhofer FIT, Schloss Birlinghoven 53757, Sankt Augustin, Germany}
\address[2]{Chair of Computer Science 5, RWTH Aachen University, Ahornstr. 55, 52074 Aachen, Germany}
\address[3]{Universidad Privada Boliviana, , Av Juan Pablo II, Cochabamba, Bolivia}

\cortext[1]{Corresponding author.}

\begin{abstract}
This study presents an approach that uses large language models such as GPT-4 to generate usage policies in the W3C Open Digital Rights Language $\mathit{ODRL}$ automatically from natural language instructions. 
Our approach uses the $\mathit{ODRL}$ ontology and its documentation as a central part of the prompt. 
Our research hypothesis is that a curated version of existing ontology documentation will better guide the policy generation. 
We present various heuristics for adapting the $\mathit{ODRL}$ ontology and its documentation to guide an end-to-end $\mathit{KG}$ construction process. 
We evaluate our approach in the context of dataspaces, i.e., distributed infrastructures for trustworthy data exchange between multiple participating organizations for the cultural domain. We created a benchmark consisting of 12 use cases of varying complexity. 
Our evaluation shows excellent results with up to 91.95\% accuracy in the resulting knowledge graph.\footnote{\href{https://github.com/Daham-Mustaf/LLM4ODRL}{LLM4ODRL: Repository containing the datasets, prompts, and code used in this paper.}}

\end{abstract}


\maketitle

\section{Introduction}
The data economy increasingly relies on dataspaces, which are distributed infrastructures for data exchange among multiple participating organizations based on data sovereignty and interoperability principles. 
To achieve interoperability in dataspaces, initiatives such as the International Data Spaces Association (IDSA)~\cite{dam2023policy} and the Gaia-X European Association for Data and Cloud strongly rely on Semantic Web technologies. 
Semantic Web technologies are not easily accessible to non-technical users – this problem also holds in dataspaces.
The specifications of both IDSA and Gaia-X rely on the W3C Open Digital Rights Language ($\mathit{ODRL}$) to describe the usage policies of assets.
Formulating usage policies in $\mathit{ODRL}$ requires familiarity with the RDF graph data model, its serializations, and the $\mathit{ODRL}$ concepts. This requirement has become a significant entry barrier with the adoption of dataspaces in highly digitalized sectors such as car manufacturing, transportation, and the cultural sector.
For example, to create the $\mathit{ODRL}$ policy from the instruction, “Painting $X$ from Museum $M$ is authorized for display within Germany but prohibited within the USA”, a domain expert is required to know all the terminology from $\mathit{ODRL}$, including \emph{Asset}, \emph{Permission}, \emph{Constraint}, etc. 

\begin{figure*}[t]
\centering
\includegraphics[width=0.6\linewidth]{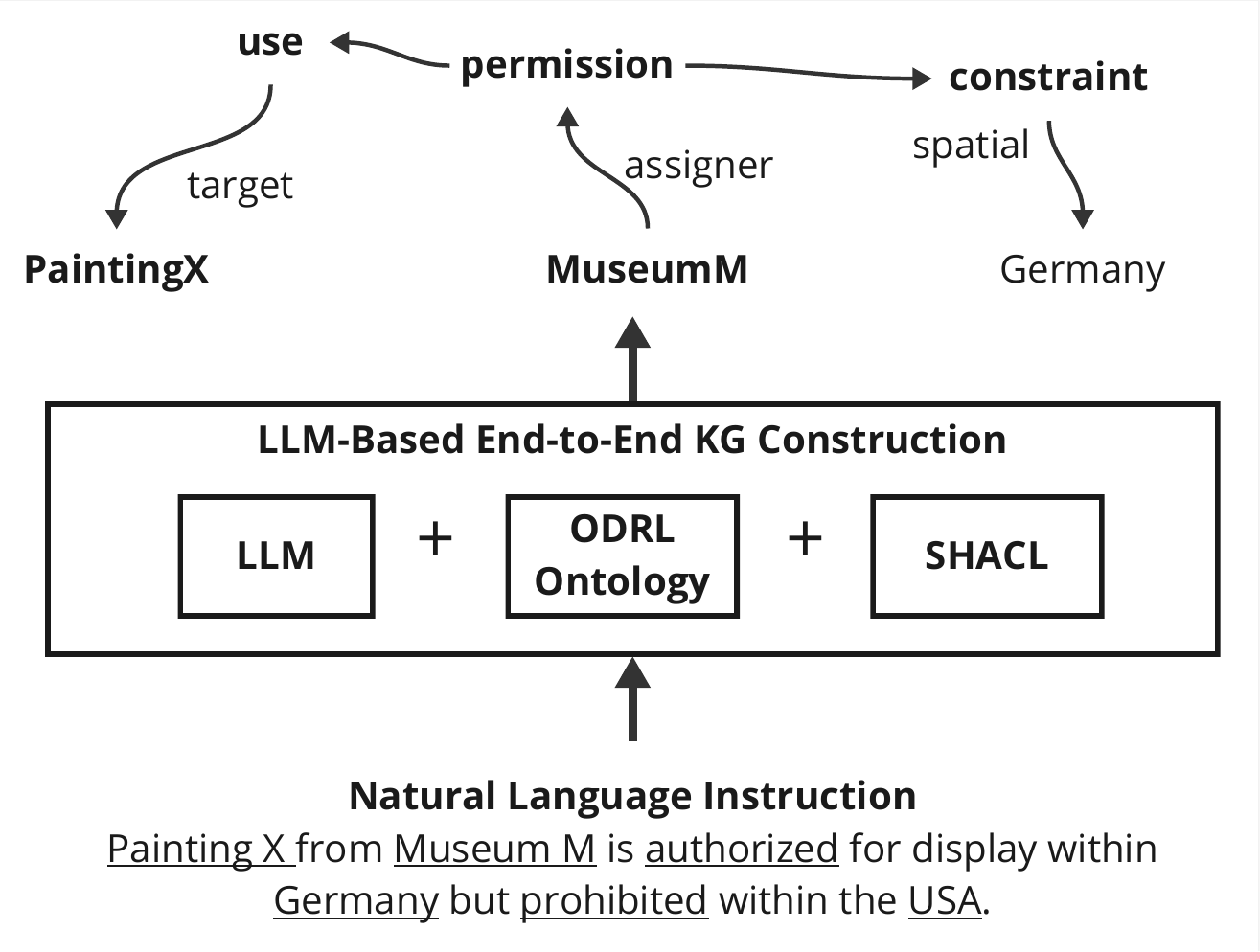}
\caption{Our method takes a natural language description of a usage policy as input. Using an LLM, a curated description of the $\mathit{ODRL}$ ontology and associated SHACL shapes, we generate a $\mathit{KG}$ that corresponds to the $\mathit{ODRL}$ representation of the described policy.}
\label{fig:motivation}
\end{figure*}

The $\mathit{ODRL}$ version 2.2, a W3C recommendation since February 2018~\cite{ODRL-Model-2018}, is an ontology for expressing rights over physical or digital goods. $\mathit{ODRL}$ enables owners and consumers to effectively express the terms and obligations governing digital asset access and use. 
We show a high-level overview of the $\mathit{ODRL}$ information model in \autoref{fig:motivation}. The main class of the Core Vocabulary is \emph{Policy}, which acts as a container for \emph{Rules}. A \emph{Set} serves as the default subclass of \emph{Policy} for conveying a \emph{Rule} over an \emph{Asset} to define general terms of usage without specific \emph{Constraint} or Duty. An \emph{Offer}, as a subclass of \emph{Policy} in the Core Model, describes the rules presented by the assigning parties and specifies the conditions for the recipient party of these rules. An \emph{Agreement} is a subclass of \emph{Policy} in the core model, which encompasses all terms governing the usage agreement between an Assigner (the party that proposes the policy statements) and an \emph{Assignee} (the party that receives the policy statements) regarding an \emph{Asset}. A \emph{Permission} specifies \emph{Actions} over an \emph{Asset}. Permissions can also be linked with a duty, especially when the action is obligatory. In contrast to permissions, \emph{Prohibitions} restrict specific actions.

To alleviate the barrier of familiarity with $\mathit{ODRL}$ and Semantic Web technologies for creating usage policies in dataspaces, we contribute with an end-to-end approach that generates usage policies in $\mathit{ODRL}$ directly from natural language instructions given by an application domain expert. 
Our approach employs Large Language Models (LLMs) combined with a hand-crafted ontology description based on the ODRL specification. 
We equip our approach with a self-correction component that evaluates the generated output and applies rules. We evaluate our approach with GPT-3.5-turbo,
GPT-4, and GPT-4o on twelve realistic use cases from the cultural domain, with increasing complexity, about the described usage permissions. We rely on SHACL (Shapes Constraint Language~\cite{SHACL-2017}) shapes to perform the evaluation from the syntax and semantic perspective.
Our results show a promising path in ontology-guided KG generation.

\section{Related Work}
Formally a knowledge graph $\mathit{KG}$ is defined as a tuple given by $\mathit{KG} = (\mathcal{E},\mathcal{R},\mathcal{T}^+)$ where $\mathcal{E}$ denotes the set of all vertices in the graph representing entities, $\mathcal{R}$ is the set of edges representing relations, and $\mathcal{T}^+ \subseteq \mathcal{E} \times \mathcal{R} \times \mathcal{E} $ is a set of all $\mathit{KG}$ triples~\cite{liu2021kg}. The task of $\mathit{KG}$ construction can be defined as, given unstructured sources $S = (s_1,s_2, \dots, s_n)$, a model $M$ being trained to approximate the function $F(S) \rightarrow \mathit{KG}$~\cite{ye-etal-2022-generative}. 
Traditionally, $\mathit{KG}$ construction comprises several tasks, namely Name Entity Recognition (NER)~\cite{NERsurvey}, Relation Extraction (RE)~\cite{xiaoyan2023comprehensive}, Entity Linking (EL)~\cite{sevgili2022neural} and Coreference Resolution (CR)~\cite{liu2023brief}. Early, the construction of knowledge graphs relied on a pipeline architecture and tools to effectively create and incrementally update a knowledge graph~\cite{hofer2023construction}. With the advent of deep learning approaches, each of these tools was further improved~\cite{zhu2023llms}. However, in recent years, Large Language Models (LLMs) have caused a paradigm shift in Natural Language Processing (NLP) by creating generalist agents~\cite{bommasani2021opportunities}. The remarkable properties of LLMs now enable end-to-end KG construction.
Early attempts included BERT style models~\cite{devlin2019google} to extract entities and associated relations~\cite{kumar2020building}, whereas Guo et al.~\cite{guo2021constructing} propose end-to-end knowledge graph construction with BERT. The work by Han et al.~\cite{han2023pive} aligns with our work, using a larger LLM (GPT-4) for KG construction. However, to correct errors in the previously generated knowledge graph, they use a small fine-tuning LLM (\emph{T5}). While more efficient during inference, this also poses the method's main drawback due to the need for labeled data that might be unavailable.

\section{Method}\label{sec:method}
\begin{figure*}[htbp!]
\centering
\includegraphics[width=\linewidth]{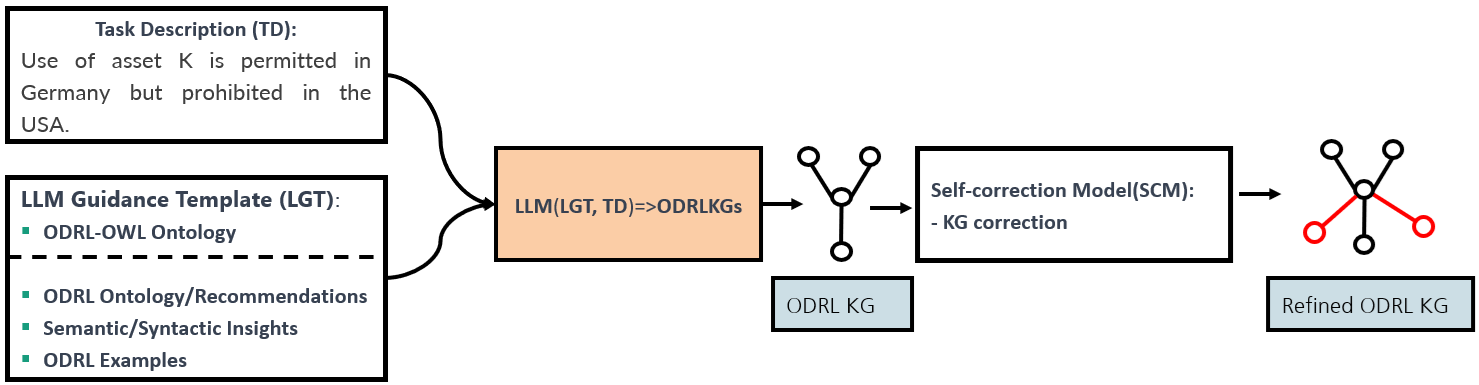}
\caption{
\textbf{Approach}. The LLM Guidance Template ($\mathit{LGT}$) comprises two modules, i.e., the $\mathit{ODRL}$ ontology in Turtle serialization and contextual insights in PDF format. $\mathit{KG}$ is constructed by passing Task Description ($\mathit{TD}$) and the $\mathit{LGT}$ as input to the LLM. Next, the Self-correction Model ($\mathit{SCM}$) refines the generated $\mathit{KG}$ by providing the TD, the first version of the $\mathit{KG}$, and the $\mathit{ODRL}$ Self-Correction Rules ($\mathit{SCR}$) as input for the LLM to produce the refined $\mathit{KG}$.
}
\label{fig:pipline}
\end{figure*}

The KG construction process begins by developing an LLM Guidance Template ($\mathit{LGT}$)\footnote{The LLM guidance templates used in this study can be found in the \href{https://github.com/Daham-Mustaf/LLM4ODRL/tree/main/llm_guidance_template/templates}{templates directory on GitHub}.}, a framework for guiding the LLM-based KG construction. We provide a Task Description ($\mathit{TD}$) in natural language with the specific requirements and constraints for the desired ODRL. The LLM then uses the $\mathit{LGT}$ and $\mathit{TD}$ to construct an $\mathit{ODRL}$ $\mathit{KG}$ end-to-end. 
After generating the $\mathit{ODRL}$ knowledge graph, our approach incorporates a Self-Correction Model ($\mathit{SCM}$) to refine the generated KG. The $\mathit{SCM}$ utilizes predefined rules to correct inconsistencies and errors within the $\mathit{KG}$, resulting in a refined $\mathit{ODRL}$ knowledge graph. 
\autoref{fig:pipline} illustrates the complete workflow of our approach, where we explain the process of generating the $\mathit{ODRL}$ and subsequently refining it using the $\mathit{SCM}$. Now, we describe each module in detail.

\subsection{Ontology-Guided KG Construction}
\label{Ontology-Guided}
From the perspective of computer science, An ontology is a formal, explicit specification of a shared conceptualization~\cite{gruber1993translation}.
The $LGT$ in this approach is based on the $\mathit{ODRL}$ version 2.2 ontology\footnote{\url{https://www.w3.org/ns/odrl/2/ODRL22.ttl}} This ontology defines $\mathit{ODRL}$ classes, core concepts, and includes various OWL axioms (e.g., domain, range, and others) that represent relationships between classes and properties. The ontology uses SKOS\footnote{\url{https://www.w3.org/2004/02/skos/}} annotations (\texttt{skos:definition}, \texttt{skos:note}, \texttt{rdfs:label}) for metadata, conceptual definitions, and labels. We pass this ontology directly to the LLM, leveraging both its formal structure and human-readable annotations. The LLM then interprets this information to generate an $\mathit{ODRL}$ KG.

\noindent Relying solely on the $\mathit{ODRL}$ ontology challenges LLMs in accurately predicting complex $\mathit{KG}$s, especially with multiple constraints (e.g., date, time, location, special conditions). LLMs may misinterpret ontology elements, leading to errors such as:
\begin{itemize}
\item Incorrect property values (e.g., using \texttt{odrl:use}, an \texttt{odrl:Action} instance, for \texttt{odrl:leftOperand}, which requires an \texttt{odrl:LeftOperand} instance)
\item Introduction of undefined properties (e.g., using \texttt{odrl:location} instead of a valid \texttt{odrl:leftOperand} value)
\end{itemize}
\noindent These errors cause inconsistencies and misalignments with the $\mathit{ODRL}$ standard. To mitigate these issues, we enrich the $\mathit{LGT}$ with additional context about $\mathit{ODRL}$ syntax and semantics, aiding LLMs in distinguishing between various data types and object properties within the ontology.

\subsection{Ontology, Syntax, Semantics, and Examples ($OSES$) Insights}
\label{subsec:ontocontext-approach}
We translate the structured $\mathit{ODRL}$ ontology into plain text, incorporating syntactic and semantic aspects to guide the LLM. However, directly importing the text of the $\mathit{ODRL}$ W3C recommendation\footnote{\url{https://www.w3.org/TR/odrl-model/}} into the $\mathit{LGT}$ file presents a significant challenge. The main issue lies in the clarity guidance. The recommendations often contain lengthy explanations and duplicated information, leading to confusion and hallucinations. In response to these challenges, we undertake a process of distillation~\cite{anand2023gpt4all}, extracting essential information from the $\mathit{ODRL}$ ontology and recommendations. As part of this distillation process, we have created guidelines inside the LGT file for the main classes of the $\mathit{ODRL}$ information model and their properties, detailing how they should be utilized by the LLM. This distilled information is then organized into three key dimensions within the LGT file, considering three main aspects: Firstly, we address $\mathit{ODRL}$'s syntactic interpretation, the syntactic structure defines how its main entities are aligned and how they should be used.
Secondly, we consider $\mathit{ODRL}$ Semantics: $\mathit{ODRL}$ \texttt{odrl:Duty} instances can represent obligations when linked to \texttt{odrl:Policy} via \texttt{odrl:obligation} or conditions for permission activation when linked to \texttt{odrl:Permission} via \texttt{odrl:duty}. Finally, integrated $\mathit{ODRL}$ examples in the LGT provide practical guidance for policy implementation, aiding in understanding and selection of relations for the output. LGT guides the LLM to generate the $\mathit{ODRL}$ $\mathit{KG}$ by solving tasks. Separate LGTs are established for each $\mathit{ODRL}$ type (\emph{Agreement}, \emph{Rule}, \emph{Offer}), providing guidance for accurate results. Despite careful design, there may be inaccuracies or discrepancies, including syntactic errors and semantic inconsistencies, in the output generated by LLM. Therefore, we apply a third self-correction model.

\subsection{$\mathit{ODRL}$ Self-Correction Model}
\label{Self-Correction Model}
For ODRL self-correction, we establish 17\footnote{\href{https://github.com/Daham-Mustaf/LLM4ODRL/blob/main/correction_report.py}{ODRL self-correction rules}} rules for \emph{Agreement}, 16 for \emph{Offer}, and $\mathit{ODRL}$ \emph{Rule}. These rules, termed correction rules, are derived from the ODRL W3C recommendation and ontology relations, ensuring alignment with the official ODRL specification and leveraging the semantic structure defined in the ontology.
$\mathit{ODRL}$ Self-Correction Rules (SCR), rules are expressed in human-readable language, which allows for accurate interpretation and application by the LLM. The inputs include the $\mathit{TD}$ along with it resulting $\mathit{ODRL}$ $\mathit{KG}$, as has been detailed previously in \cref{Ontology-Guided}, and the SCR. These inputs are then passed to the LLM. We have two options for applying self-correction. The first option is to define each rule separately and iterate over them. The second option is to consolidate all rules into one prompt and then pass it to the LLM. Although this method is less expensive than the first option, it may slightly compromise accuracy due to the large text processing required by the LLM. The SHACL violation messages alone do not effectively guide the LLM to produce KG correction. This is due to the stateless nature of LLMs, which do not retain memory of previous interactions or learn from individual correction attempts. As such, simply passing SHACL violation messages back to the LLM does not guarantee improved KG.
The LLM compares the $\mathit{ODRL}$ $KG$ policy with the $SCR$. If any inconsistencies are detected, the LLM makes adjustments to ensure compliance with the $SCR$. This results in a refined $\mathit{ODRL}$ $KG$, as illustrated in \autoref{fig:pipline}. Unlike SHACL shapes, which primarily detect violations without modifying the $KG$, the LLM can autonomously refine the KG based on detected rule violations.

\section{Evaluation}\label{sec:evaluation}
Evaluating outputs from LLMs is crucial due to their probabilistic nature. Unlike deterministic systems, LLMs may generate varying outputs for the same task, as LLMs may still generate probabilistic outputs even when provided with factual data.
We conducted a comprehensive set of 108 experiments to evaluate the performance of three different LLM models: GPT-3.5-turbo, GPT-4, and GPT-4o. For each use case, we employed three different methodologies: Ontology-Guided, OSES Insights, and Refinement, resulting in 36 experiments per model. The experiments aimed to assess the models' capabilities in managing and enforcing digital rights.

\textbf{Dataset:}
Inspired by the context of the project Datenraum Kultur(DRK)\footnote{\url{https://datenraum-kultur.fit.fraunhofer.de/}}, We designed a dataset of twelve different use cases.\footnote{The use cases can be found in the \href{https://github.com/Daham-Mustaf/LLM4ODRL/blob/main/data/tasks/use_cases.yaml}{YAML file on GitHub}.} These use cases represent tasks defined in plain text, designed to prompt LLMs to generate ODRL KGs. These use cases are derived from real scenarios in the context of the DRK and encapsulate different criteria for determining the appropriate application of policies to assets where digital rights need to be managed and enforced in the cultural domain. Each use case serves as a task description for the LLM. The dataset includes 4 \emph{Agreement}, 5 \emph{Offer} and 3 \emph{Set} of type $\mathit{ODRL}$ \emph{Policy}. Case 1, illustrated below, demonstrates a typical scenario for ODRL policy generation. This example policy regulates access to the ShowTimesAPI' dataAPI managed by DE\_Staatstheater\_Augsburg, a German cultural organization. The policy's main objective is to control access to valuable cultural assets stored in the dataAPI. Access is restricted to subscribers such as Regional Newspaper', Culture Research Institute', and Cultural Platform Bavaria'. Furthermore, access is limited to users located in Germany, and usage rights expire on May 10, 2025.


\textbf{Criteria Definition and Constraint Establishment:}
Criteria $C_1$--$C_9$~\autoref{tab:odrl-criteria} are defined for each use case based on the W3C $\mathit{ODRL}$ recommendation, focusing on both semantic and syntactic representations. Each criterion sets expectations for an $\mathit{ODRL}$ representation, ensuring that the selected use cases comply with both the structural and content constraints necessary for effective digital rights management. The next step is mapping to $\mathit{SHACL}$ constraints for validation. Each criterion is translated into a $\mathit{SHACL}$  shape with associated constraints manually generated for this study\footnote{\href{https://github.com/Daham-Mustaf/LLM4ODRL/tree/main/ODRL_policy_validation_shapes}{SHACL shapes for ODRL KGs evaluation}}. This constraint encompasses five shapes: \texttt{PolicyShape}, \texttt{PermissionShape}, \texttt{PartyShape}, \texttt{AssetShape}, and \texttt{ConstraintShape} collectively evaluate 22 ranges of properties for \texttt{Agreement}, 20 for \texttt{Offer}, and 16 for \texttt{Asset} policy types.
\begin{table}[h]
    \centering
    \caption{$\mathit{ODRL}$ Criteria and Violations}
     \label{tab:odrl-criteria} 
    \begin{tabular}{|p{2cm}|p{8cm}|p{2.35cm}|}
    \hline
    \textbf{Name} & \textbf{Assertion} & \textbf{SHACL} \\
    \hline
    $C_1$: odrl:uid & Instances of classes \texttt{Policy}, \texttt{Asset}, \texttt{Party}, and \texttt{Constraint} \textbf{MUST} be associated with an (\texttt{odrl:uid}). & PolicyShape. \\
    \hline
    $C_2$: Data type specification & \textbf{MUST} explicitly specify the corresponding XML Schema Definition (XSD) data type. & All Shapes\\
    \hline
    $C_3$: Meta info & \textbf{MUST} include mandatory meta-information: \texttt{dc:creator}, \texttt{dc:title}, \texttt{dc:description}, and \texttt{dc:issued}. & PolicyShape \\
    \hline
    $C_4$: Function & \texttt{Offer} \textbf{MUST} have one \texttt{odrl:assigner} of type \texttt{Party}. \texttt{Agreement} \textbf{MUST} have one \texttt{odrl:assigner} and \texttt{odrl:assignee} of type \texttt{Party}.& PartyShape \\
    \hline
    $C_5$: Relation & Policy \textbf{MUST} have one \texttt{odrl:target} property with an object of type \texttt{Asset}.& AssetShape \\
    \hline
    $C_6$: Action & Policy\textbf{MUST} have one \texttt{odrl:action} property of type \texttt{Action}.
 & PermissionShape \\
    \hline
    $C_7$: Rule &Policy \textbf{MUST} contain at least one rule specifying actions on \texttt{Asset}s, which can be \texttt{Permission}, \texttt{Prohibition}, or \texttt{Obligation}.
 & PolicyShape \\
    \hline
    $C_8$: Constraint &\texttt{Rule} Constraint \textbf{MUST} have properties \texttt{leftOperand} (\texttt{LeftOperand}), \texttt{operator} of type (\texttt{Operator}), and \texttt{rightOperand} (Literal, IRI, or \texttt{RightOperand}).
 & ConstraintShape \\
    \hline
    $C_9$: $\mathit{ODRL}$ Extension &New elements introduced by LLM \textbf{MUST} either be explicitly defined in $\mathit{ODRL}$ or adhere to the the $\mathit{ODRL}$ Profile Mechanism. \footnote{\href{https://www.w3.org/TR/odrl-model/\#profile-mechanism}{$\mathit{ODRL}$ Profile Mechanism}}.
 & All shapes \\
    \hline
    \end{tabular}
\end{table}

\textbf{Assignment of Scores:}
Each Focus Nodes and property shape in side data graph evaluated by SHACL shapes is assigned a score of 1 if it meets the defined criteria and 0 if not. We utilized the \emph{SHACL Playground}\footnote{SHACL Playground: \url{https://shacl.org/playground/}} for determining the scores. 
In addition, a script was used to automate the calculation of these scores. After evaluating the scores for each use case, we aggregated the results to provide an overview of the performance across all scenarios for each approach developed in our methodology. These scores, representing metrics for the quality of $\mathit{KG}$ use cases, serve as accuracy indicators. Higher scores directly correspond to greater accuracy and better adherence to the defined criteria, reflecting a more precise and effective representation of the use cases.

\begin{figure*}[t]
\centering
\includegraphics[width=\linewidth]{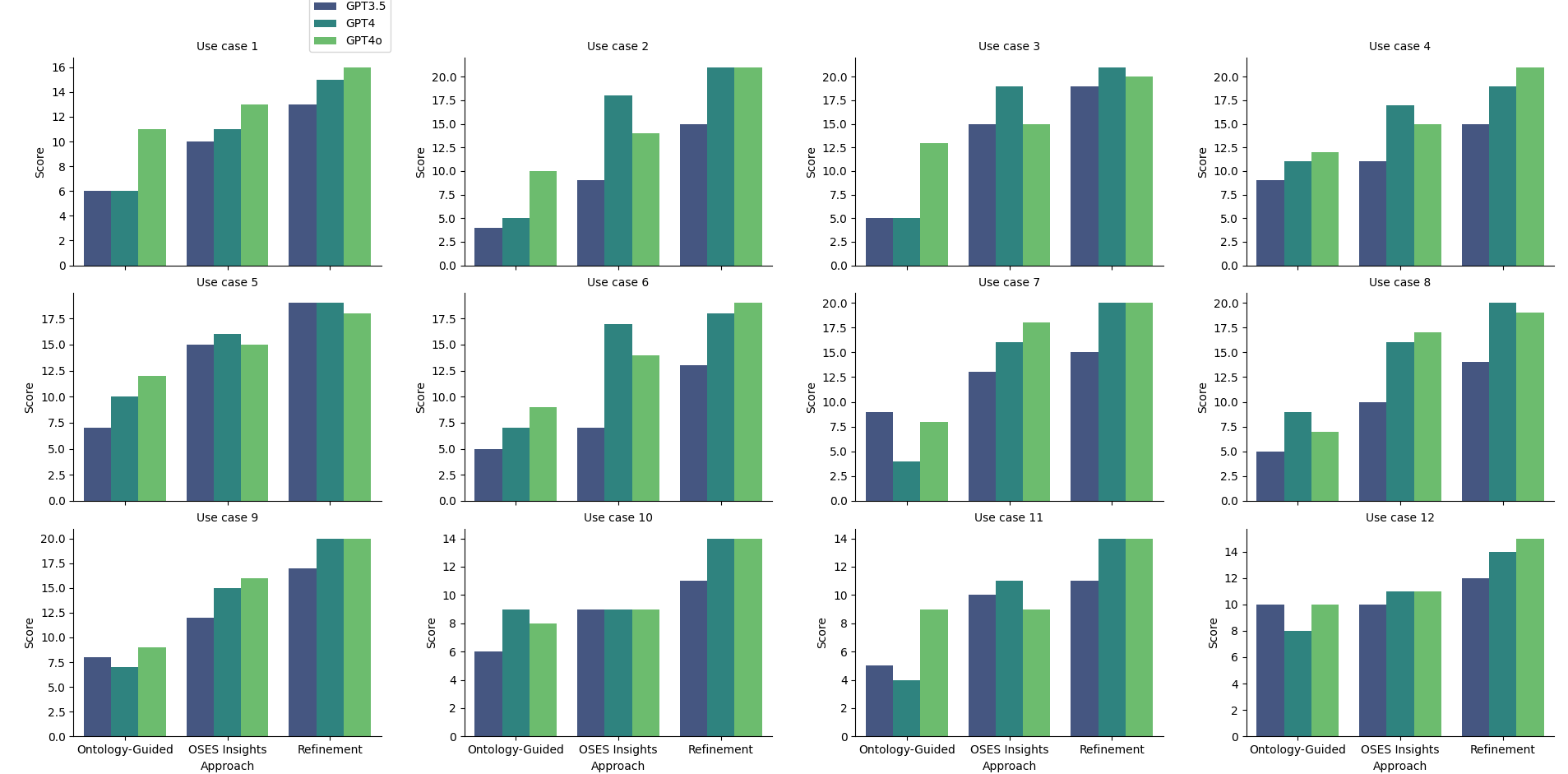}
\caption{Figure 3 illustrates performance across 12 use cases for GPT-3.5-turbo, GPT-4, and GPT-4o. LLMs learn progressively from ontology input, with improved context and examples enhancing performance. The Refinement method, combining ontology guidance and self-correction, consistently achieves the highest scores across all models and use cases, demonstrating its effectiveness in generating accurate ODRL policies.}
\label{fig:scores}
\end{figure*}

\textbf{Result:}
In the Ontology-Guided approach, the LLM faced challenges in extracting information solely from $\mathit{ODRL}$ ontology. The large size compounded these challenges. Without the assistance of real $\mathit{ODRL}$ examples, the LLM struggled to accurately predict the $\mathit{ODRL}$ Figure~\ref{fig:scores}. However, in the $OSES$ Insights approach, the $\mathit{ODRL}$ $KG$ is enhanced with properties and relations. By formulating the ontology in text and providing semantic, syntactic, and real examples, the guidance provided to the LLM is improved. Refined ODRL KG quality improves through self-correction rules\footnote{\url{https://github.com/Daham-Mustaf/LLM4ODRL/blob/main/correction_report.py}} targeting crucial omitted properties. This iterative process enhances KG completeness and accuracy, with effectiveness highly dependent on the initial KG generation quality. 
Initially, we ran each use case once and scored the output as the first exploration. the results indicate varying degrees of performance across the LLM variants and methodologies, with GPT-4 and GPT-4o exhibiting similar performance patterns across the use cases. 


\[ \text{Accuracy} = \frac{\sum R}{\sum T} \times 100\% \]
\scriptsize
\begin{tabular}{@{}l@{\hspace{0.5em}}l@{}}
$R$: Total obtained score & $T$: Total possible score \\[0.5ex]
\multicolumn{2}{@{}l}{Use Case 1 (with refinement):} \\
\multicolumn{2}{@{}l}{$\frac{217}{236} \times 100\% \approx 91.95\%$}
\end{tabular}
\normalsize

\section{Conclusions}
This work proposes a novel methodology for generating KGs from textual data using an LLM. We demonstrate that interpreting the ontology in plain text significantly boosts LLM accuracy. The LLM can compare KGs with textual predefined rules presented in human-readable text and refine the results. This capability to refine KGs underscores the advantages of employing LLM in KG construction.
One limitation of our current work is that it is evaluated on using the $\mathit{ODRL}$ ontology. However, our approach is generic and can be extended to diverse ontologies.
In our current methodology, $LGT$ guidelines for KG construction (section \ref{Ontology-Guided}) and selfcorrection rules (section \ref{Self-Correction Model}) have been derived and formulated from the W3C Recommendation  and $\mathit{ODRL}$ ontology, which involves manual interpretation and analysis. A potential area for future work could be to automate this task, e.g., by identifying and extracting the most important parts of the ontology using a statistical analysis of existing knowledge graphs that are built upon the ontology. An LLM could be used to verbalize the remaining parts.

\begin{acknowledgments}
This work was supported by the German Ministry for Research and Education (BMBF) project WestAI (Grant no. 01IS22094D) and the German Federal Commissioner for Culture and the Media via Datenraum Kultur (Grant no. 2522DIG012).
\end{acknowledgments}

\bibliography{bibliography-ceur}

\end{document}